\documentclass{article}

\usepackage{arxiv}

\usepackage[utf8]{inputenc} 
\usepackage[T1]{fontenc}    
\usepackage{hyperref}       
\usepackage{url}            
\usepackage{booktabs}       
\usepackage{amsfonts}       
\usepackage{nicefrac}       
\usepackage{microtype}      
\usepackage{tikz}
\usepackage{amsmath,amssymb}
\usepackage{subfig}

\newtheorem{definition}{Definition}

\newcommand{\problem}[1]{{\sc \bf #1}}
\newcommand{\CE}{\problem{CE}}
\newcommand{\EE}{\problem{EE}}
\newcommand{\SE}{\problem{SE}}
\newcommand{\DS}{\problem{DS}}
\newcommand{\DC}{\problem{DC}}
\newcommand{\semantics}[1]{{\sc \bf #1}}
\newcommand{\CF}{\semantics{CF}}
\newcommand{\ADM}{\semantics{ADM}}
\newcommand{\CO}{\semantics{CO}}
\newcommand{\PR}{\semantics{PR}}
\newcommand{\ST}{\semantics{ST}}
\newcommand{\SST}{\semantics{SST}}
\newcommand{\STG}{\semantics{STG}}
\newcommand{\ID}{\semantics{ID}}

\title{Design and Results of ICCMA 2021}

\author{
  Jean-Marie Lagniez \\
  CRIL, University of Artois and CNRS\\
  \texttt{lagniez@cril.fr} \\
   \And
 Emmanuel Lonca \\
  CRIL, University of Artois and CNRS\\
  \texttt{lonca@cril.fr} \\
  \AND
  Jean-Guy Mailly \\
  LIPADE, University of Paris \\
  \texttt{jean-guy.mailly@u-paris.fr} \\
  \And
  Julien Rossit \\
  LIPADE, University of Paris \\
  \texttt{julien.rossit@u-paris.fr} \\
}

\begin{document}
\maketitle

\begin{abstract}
Since 2015, the International Competition on Computational Models of Argumentation (ICCMA) provides a systematic comparison of the different algorithms for solving some classical reasoning problems in the domain of abstract argumentation. This paper discusses the design of the Fourth International Competition on Computational Models of Argumentation. We describe the rules of the competition and the benchmark selection method that we used. After a brief presentation of the competitors, we give an overview of the results.
\end{abstract}

Formal argumentation \cite{HOFA,HOFA2} is a major topic in the domain of knowledge representation and reasoning. This formalism allows to reason with conflicting information, and has applications {\em e.g.} in automated negotiation \cite{DimopoulosMM21} or decision making \cite{AmgoudV12}. The most classical reasoning tasks in this kind of formalism are intractable in the general case (see {\em e.g.} \cite{HOFAComplexity} for an overview of computational complexity in formal argumentation). This has conducted to the organization of the International Competition on Computational Models of Argumentation (ICCMA), that allows to compare the efficiency of the different algorithms that have been proposed for these reasoning problems. The previous competitions \cite{ThimmV17,GagglLMW20,BistarelliK0T20} have shown that, in spite of the theoretical hardness of argumentative reasoning, some powerful techniques allow to handle them efficiently.

After a short introduction to abstract argumentation in Section~\ref{section:background}, we describe the rules of the competition in Section~\ref{section:rules}. In particular, we describe the various (sub)tracks as well as the scoring system and the benchmark selection. Note that, contrary to what was initially announced \cite{LagniezLMR20}, there has been no track dedicated to dynamic argumentation or structured argumentation at ICCMA 2021. However, there has been some interest in the community that conducted us to add a new track dedicated to approximation algorithms for reasoning with abstract argumentation frameworks. Then, we present the competitors and the results in Section~\ref{section:participants-and-results}.

\section{Background: Abstract Argumentation}\label{section:background}
An abstract argumentation framework (AF) \cite{Dung95} is a directed graph $F = \langle A, R\rangle$, where $A$ is the set of arguments, and $R \subseteq A \times A$ is the attack relation. For $a,b,c\in A$, we say that $a$ {\em attacks} $b$ if $(a,b) \in R$. If in turn $b$ attacks $c$, then $a$ {\em defends} $c$ against $b$. Similarly, a set $S \subseteq A$ attacks (respectively defends) an argument $b$ if there is some $a \in S$ that attacks (respectively defends) $b$. For $S \subseteq A$ a set of arguments, $S^+$ is the set of arguments that are attacked by $S$, formally $S^+ = \{b \in A \mid \exists a \in S$ s.t. $(a,b) \in R\}$. The range of $S$ is $S^\oplus = S \cup S^+$.

Different semantics have been defined for evaluating the acceptability
of (sets of) arguments.

\begin{definition}\label{definition:semantics}
  Given an AF $F = \langle A, R\rangle$, a set of arguments $S \subseteq A$ is conflict-free iff $\forall a,b \in S$, $(a,b) \not\in R$. A conflict-free set $S$ is admissible iff $\forall a \in S$, $S$ defends $a$ against all its attackers. Conflict-free and admissible sets are respectively denoted by $\CF(F)$ and $\ADM(F)$.

  Now, we formally introduce the extension-based semantics. For $S \subseteq A$,
  \begin{itemize}
    \item $S \in \CO(F)$ iff $S \in \ADM(F)$ and $\forall a \in A$ that is defended by $S$, $a \in S$;
    \item $S \in \PR(F)$ iff $S$ is a $\subseteq$-maximal admissible set;
    \item $S \in \ST(F)$ iff $S$ is a conflict-free set that attacks each $a \in A \setminus S$;
    \item $S \in \SST(F)$ iff $S \in \CO(F)$ and there is no $S_2 \in \CO(F)$ s.t. $S^\oplus \subset S_2^\oplus$; 
    \item $S \in \STG(F)$ iff $S \in \CF(F)$ and there is no $S_2 \in \CF(F)$ s.t. $S^\oplus \subset S_2^\oplus$; 
    \item $S \in \ID(F)$ iff $S \in \ADM(F)$, $S \subseteq \cap \PR(F)$, and there is no $S_2 \subseteq \cap \PR(F)$ such that $S_2 \in \ADM(F)$ and $S \subset S_2$.
  \end{itemize}
\end{definition}
\CO, \PR, \ST, \SST, \STG{} and \ID{} stand (respectively) for the
complete, preferred, stable \cite{Dung95}, semi-stable \cite{CaminadaCD12}, stage
\cite{Verheij96} and ideal \cite{DungMT07} semantics. We refer the
interested reader to \cite{HOFASemantics} for more details about these semantics.

For $\sigma \in \{\CO,\PR,\ST,\SST,\STG,\ID\}$ a semantics, an
argument $a \in A$ is credulously (respectively skeptically) accepted in $F =
\langle A, R\rangle$ with respect to $\sigma$ iff $a \in S$ for some
(respectively each) $S \in \sigma(F)$.

We are interested in various reasoning problems:
\begin{itemize}
\item \CE-$\sigma$: Given an AF $F = \langle A, R\rangle$, give the
  number of $\sigma$-extensions of $F$.
\item \SE-$\sigma$: Given an AF $F = \langle A, R\rangle$, give one
  $\sigma$-extension of $F$.
\item \DC-$\sigma$: Given an AF $F = \langle A, R\rangle$ and $a \in
  A$ an argument, is $a$ credulously accepted in $F$?
\item \DS-$\sigma$: Given an AF $F = \langle A, R\rangle$ and $a \in
  A$ an argument, is $a$ skeptically accepted in $F$?
\end{itemize}
Most of these problems are computationally hard in general, under the semantics from Definition~\ref{definition:semantics}. For instance, the decision problems \DC{} and \DS{} might be complete for the first or second level of the polynomial hierarchy (see \cite{HOFAComplexity} for more details).

\section{Rules of the Competition}\label{section:rules}
\subsection{Tracks and Subtracks}
The competition is made of two main tracks. The first one, the most ``classical" one, is dedicated to exact algorithms for reasoning with abstract AFs. The second track has been introduced for the first time at ICCMA 2021, and is dedicated to approximation algorithms for abstract argumentation.

\paragraph{Exact Algorithms} The first track is made of six subtracks, each of them corresponding to one of the six semantics under consideration. Each subtrack is made of several reasoning tasks. More precisely, for $\sigma \in \{\CO, \PR, \ST, \SST, \STG\}$, all the problems \CE-$\sigma$, \SE-$\sigma$, \DC-$\sigma$ and \DS-$\sigma$ must be solved. For $\sigma = \ID$, since any AF possesses exactly one ideal extension, \CE-$\sigma$ is trivial, and \DC-$\sigma$ is equivalent to \DS-$\sigma$. So, only \SE-$\sigma$ and \DS-$\sigma$ are considered.

\paragraph{Approximate Algorithms} The second track is also made of six subtracks corresponding to the six semantics. However, the problems \CE-$\sigma$ and \SE-$\sigma$ are not included in the subtracks this time, {\em i.e.} only the decision problems \DC-$\sigma$ and \DS-$\sigma$ appear. This means that every subtrack is made of two problems, except $\sigma = \ID{}$ which is made of only one task.

\subsection{Input/Output and Environment}
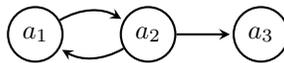
\begin{figure}[ht]
\centering
  \begin{tikzpicture}[->,>=stealth,shorten >=1pt,auto,node distance=1.5cm, thick,main node/.style={circle,draw,font=\bfseries}]
    \node[main node] (a1) {$a_1$};
    \node[main node] (a2) [right of=a1] {$a_2$};
    \node[main node] (a3) [right of=a2] {$a_3$};
    
    \path[->] (a1) edge[bend left] (a2)
    (a2) edge (a3)
    (a2) edge[bend left] (a1);
  \end{tikzpicture}
\caption{The AF $F$\label{fig:example-af-formats}}
\end{figure}

The input/output formats are the same as the formats from ICCMA 2019 for all the reasoning tasks that were considered then; only \CE-$\sigma$ requires the definition of a new format. Each benchmark is provided in two different file formats: trivial graph format (\verb+tgf+) and ASPARTIX format (\verb+apx+). For the following examples, we use the AF $F = \langle A, R\rangle$ with $A = \{a_1,a_2,a_3\}$ and $R = \{(a_1,a_2), (a_2,a_3), (a_2,a_1)\}$, depicted at Figure~\ref{fig:example-af-formats}.

The Trivial Graph Format describes a graph by giving a list of identifiers for nodes, then a list of edges, separated by a $\#$ symbol. See \url{https://en.wikipedia.org/wiki/Trivial_Graph_Format} for more details. We give below the content of \verb+myFile.tgf+ that corresponds to the AF depicted at Figure~\ref{fig:example-af-formats}.

\begin{verbatim}
1
2
3
#
1 2
2 3
2 1
\end{verbatim}

The ASPARTIX format (named after the ASP-based argumentation solver ASPARTIX \cite{EglyGW10}) describes the argument names as rules \verb+arg(name).+, and attacks as rules \verb+att(name1,name2).+ We give below, as an example, the content of \verb+myFile.apx+ that corresponds to the AF depicted at Figure~\ref{fig:example-af-formats}.

\begin{verbatim}
arg(a1).
arg(a2).
arg(a3).
att(a1,a2).
att(a2,a3).
att(a2,a1).
\end{verbatim}

For both tracks, solvers must write the result to standard output exactly in the format described below.
\begin{itemize}
	\item \DC-$\sigma$ and \DS-$\sigma$, for $\sigma \in \{\CO,\PR,\ST,\SST,\STG,\ID\}$. The output must be either
	\begin{verbatim}
	YES
	\end{verbatim}
	if the queried argument is (respectively) credulously or skeptically accepted in the given AF under $\sigma$, or
	\begin{verbatim}
	NO
	\end{verbatim}
	otherwise.
	\item \SE-$\sigma$, for $\sigma \in \{\CO,\PR,\ST,\SST,\STG,\ID\}$. The output must be of the form
	\begin{verbatim}
	[a1,a2,a3]
	\end{verbatim}
	meaning that $\{a_1,a_2,a_3\}$ is a $\sigma$-extension of the given AF, where $a_1, a_2$ and $a_3$ are arguments. If $\sigma = \ST$, there may be benchmarks that do not possess any extension. In that case, the output must be
	\begin{verbatim}
	NO
	\end{verbatim}
	\item \CE-$\sigma$, for $\sigma \in \{\CO,\PR,\ST,\SST,\STG\}$. The output must be of the form
	\begin{verbatim}
	k
	\end{verbatim}
	where $k \in \mathbb{N}$ is the number of $\sigma$-extensions of the given AF.
\end{itemize}

The solver interface is also inspired from ICCMA 2019. The new problem \CE{} has a similar command line to the previous \EE{} problem.\footnote{See \url{http://argumentationcompetition.org/2021/SolverRequirements.pdf} for more details.}

The competition has been run on a computer cluster where each machine has an Intel Xeon E5-2637 v4 CPU and $128$GB of RAM. The runtime limit for each instance is $600$ seconds for the ``exact" track, and $60$ seconds for the ``approximate" track. The memory limit is the machine's memory, {\em i.e.} $128$GB.

\subsection{Scoring Rules}
There is one ranking for each sub-track, {\em i.e.} six rankings for the ``exact" track and six rankings for the ``approximate" track. To be ranked, a solver must participate to the full subtrack (but without any obligation to participate to all the (sub)tracks). The scoring system is slightly different between both tracks.

For the ``exact" track, any wrong result on an instance $i$ in a subtrack conducts to the exclusion of the solver from the said subtrack. It does not prevent the solver from being ranked for other subtracks if there is no wrong results for these other ones. Then, the score of a solver $\mathcal{S}$ on the instance $i$ is
\[
Score(\mathcal{S},i) = \left \{
\begin{array}{lr}
	1 & \text{the correct answer is given in the runtime limit}\\
	0 & \text{timeout or non-parsable output}
\end{array}
\right .
\]

On the contrary, wrong results do not lead to an exclusion in the ``approximate" track:
\[
Score(\mathcal{S},i) = \left \{
\begin{array}{lr}
	1 & \text{the correct answer is given in the runtime limit}\\
	0 & \text{wrong result, timeout or non-parsable output}
\end{array}
\right .
\]

Then, the score of the solver $\mathcal{S}$ for the task $\mathcal{T}$ is
\[
Score(\mathcal{S},\mathcal{T}) = \sum_{i \in \mathcal{T}} Score(\mathcal{S},i)
\]
and the score for the subtrack $\mathcal{ST}$ is
\[
Score(\mathcal{S},\mathcal{ST}) = \sum_{T \in \mathcal{ST}} Score(\mathcal{S},\mathcal{T}).
\]

In the case where two solvers have the same score for a given subtrack, the cumulated runtime over the instances correctly solved is used as a tie-break rule (the fastest is the best).

\subsection{Benchmark Selection}
\paragraph{ICCMA 2019 Instances} We have selected $165$ instances from the previous competition. The goal is to observe the evolution of the algorithmic techniques for abstract argumentation during the last two years. These instances are the hardest ones from ICCMA 2019, with respect to two criteria:
\begin{itemize}
	\item the number of times some solvers have reached the timeout on these instances;
	\item the average runtime for solving these instances.
\end{itemize}

\paragraph{New instances} We have defined a new method for generating challenging benchmarks. This method has been used for generating $422$ new instances. For creating an instance, we use this procedure:
\begin{enumerate}
	\item Generate a (meta-)graph $G$ following a classical generation pattern ({\em e.g.} Erdos-Renyi, Barabasi-Albert,$\dots$).
	\item For each node $n_i$ in this graph, generate a new graph $F_i$.
	\item For each edge $(n_1,n_2)$ in $G$, pick some arguments $a_1$ in $F_1$ and $a_2$ in $F_2$, and add an edge $(a_1,a_2)$.
\end{enumerate}
Intuitively, this method is used to create graphs with ``communities of arguments".

\paragraph{Query argument selection} The last part of the benchmark selection is the choice of an argument to be queried for skeptical or credulous acceptance (\DS, \DC). Simply, for each AF, one argument is randomly chosen, and this argument is used for all the \DS{} and \DC{} queries on the same AF.

\section{Participants and Results}\label{section:participants-and-results}
\subsection{Competitors}
There were $9$ solvers participating to ICCMA 2021, $7$ in the exact track, and $2$ in the new approximate track.

\paragraph{Exact solvers}
\begin{itemize}
	\item A-Folio DPDB (Fichte, Hecher, Gorczyca and Dewoprabowo) \cite{A-Folio-DPDB}
	\item ASPARTIX-V21 (Dvor\'ak, K\"onig, Wallner and Woltran) \cite{ASPARTIX}
	\item ConArg (Bistarelli, Rossi, Santini and Taticchi) \cite{ConArg}
	\item FUDGE (Thimm, Cerutti, Vallati) \cite{Fudge}
	\item MatrixX (Heinrich) \cite{MatrixX}
	\item $\mu$-toksia (Niskanen and J\"arvisalo) \cite{mu-toksia}
	\item PYGLAF (Alviano) \cite{Pyglaf}
\end{itemize}

Observe that $3$ solvers are new submissions (A-Folio DPDB, FUDGE and MatrixX) , while $4$ of them (ASPARTIX-V21, ConArg, $\mu$-toksia and PYGLAF) are updated versions of solvers submitted to previous editions of ICCMA. Various techniques have been used, like translations into SAT, Constraint Programming or ASP, and dedicated algorithms.

\paragraph{Approximate solvers}
\begin{itemize}
	\item AFGCN (Malmqvist) \cite{AFGCN}
	\item HARPER$++$ (Thimm) \cite{Harper}
\end{itemize}

AFGCN is mainly based on neural networks, while HARPER$++$ uses the grounded semantics as a tool for approximating the results of the various decision problems.

\paragraph{Participation to Subtracks} The solvers are registered to ICMMA 2021 (sub)tracks as described by Table~\ref{table:participation-solvers}.
\begin{table}[htb]
\centering
\begin{tabular}{c|cccccc|cccccc}
	\hline
	& \multicolumn{6}{c}{Exact Track} & \multicolumn{6}{|c}{Approximate Track} \\
	Solver & \CO & \PR & \ST & \SST & \STG & \ID & \CO & \PR & \ST & \SST & \STG & \ID \\ \hline
	AFGCN & & & & & & & $\checkmark$ & $\checkmark$ & $\checkmark$ & $\checkmark$ & $\checkmark$ & $\checkmark$ \\
	A-Folio DPDB & $\checkmark$ & & $\checkmark$ & & & & & & & & & \\
	ASPARTIX-V21 & $\checkmark$ & $\checkmark$ & $\checkmark$ & $\checkmark$ & $\checkmark$ & $\checkmark$ & & & & & & \\
	ConArg & $\checkmark$ & $\checkmark$ & $\checkmark$ & $\checkmark$ & $\checkmark$ & $\checkmark$ & & & & & & \\
	FUDGE & $\checkmark$ & $\checkmark$ & $\checkmark$ & & & $\checkmark$ & & & & & & \\
	HARPER$++$ & & & & & & & $\checkmark$ & $\checkmark$ & $\checkmark$ & $\checkmark$ & $\checkmark$ & $\checkmark$ \\
	MatrixX & $\checkmark$ & & $\checkmark$ & & & & & & & & & \\
	$\mu$-toksia & $\checkmark$ & $\checkmark$ & $\checkmark$ & $\checkmark$ & $\checkmark$ & $\checkmark$ & & & & & & \\
	PYGLAF & $\checkmark$ & $\checkmark$ & $\checkmark$ & $\checkmark$ & $\checkmark$ & $\checkmark$ & & & & & & \\ \hline
\end{tabular}
\caption{Participation of the solvers to the various subtracks \label{table:participation-solvers}}
\end{table}

\subsection{Results of the Exact Track}
Results for the track dedicated to exact algorithms are described in Table~\ref{tab:ranking-exact}. Half of the subtracks are won by new solvers (A-Folio-DPDB for the complete and stable semantics, and Fudge for the ideal semantics), while updated versions of existing solvers perform well in the other subtracks (PYGLAF for the preferred and semi-stable semantics, and ASPARTIX for the stage semantics).
An interesting point is the global correctness of the solvers. Only two solvers presented minor bugs during the running of the competition (Fudge on the stable semantics, and PYGLAF on the stage semantics). We gave some feedback to the competitors, which allowed to correct the mistakes, and only PYGLAF had to be excluded from one subtrack (the one dedicated to the stage semantics).
Detailed results (ranking for each subtrack, with the cumulated runtime over the successfully solved instances) are given in Appendix~\ref{appendix:exact}.

\begin{table}[htb]
\centering
\subfloat[Complete Semantics\label{tab:exact-complete}]{
\begin{tabular}{ccc}
	\hline
	Rank & Solver & Score \\ \hline
	1 & A-Folio DPDB &1838 \\
	2 & PYGLAF &1835 \\
	3 & $\mu$-toksia &1803 \\
	4 & ASPARTIX-V21 & 1787 \\
	5 & FUDGE & 1695 \\
	6 & MatrixX & 759 \\
	7 & ConArg & 428 \\
	\hline
\end{tabular}
}
\subfloat[Stable Semantics\label{tab:exact-stable}]{
\begin{tabular}{ccc}
	\hline
	Rank & Solver & Score \\ \hline
	1 & A-Folio-DPDB & 1862 \\
	2 & PYGLAF & 1743 \\
	3 & FUDGE & 1585 \\
	4 & $\mu$-toksia & 1441 \\
	5 & ASPARTIX-V21 & 1429 \\
	6 & ConArg & 429 \\
	7 & MatrixX & 259 \\
	\hline
\end{tabular}
}

\subfloat[Preferred Semantics\label{tab:exact-preferred}]{
\begin{tabular}{ccc}
	\hline
	Rank & Solver & Score \\ \hline
	1 & PYGLAF& 1299 \\
	2 & $\mu$-toksia & 1210 \\
	3 & FUDGE & 1190 \\
	4 & ASPARTIX-V21 & 1052 \\
	5 & ConArg & 429 \\ 
	\hline
\end{tabular}
}
\subfloat[Ideal Semantics\label{tab:exact-ideal}]{
\begin{tabular}{ccc}
	\hline
	Rank & Solver & Score \\ \hline
	1 & FUDGE & 492 \\
	2 & ASPARTIX-V21 & 306 \\
	3 &  PYGLAF & 238 \\
	4 & $\mu$-toksia & 216 \\
	5 & ConArg & 214 \\
	\hline
\end{tabular}
}

\subfloat[Semi-Stable Semantics\label{tab:exact-semi-stable}]{
\begin{tabular}{ccc}
	\hline
	Rank & Solver & Score \\ \hline
	1 & PYGLAF & 1515 \\
	2 & $\mu$-toksia & 1103 \\
	3 & ASPARTIX-V21 &  744 \\
	4 & ConArg & 428 \\
	\hline
\end{tabular}
}
\subfloat[Stage Semantics\label{tab:exact-stage}]{
\begin{tabular}{ccc}
	\hline
	Rank & Solver & Score \\ \hline
	1 & ASPARTIX-V21 & 879 \\
	2 & $\mu$-toksia & 788 \\
	3 & ConArg & 425 \\
	\hline
\end{tabular}
}
\caption{Rankings for the {\em Exact track} \label{tab:ranking-exact}}
\end{table}

\subsection{Results of the Approximate Track}
The results for the various subtracks of the Approximate Track are described in Table~\ref{tab:ranking-approximate}. AFGCN wins most of the subtracks, but it is dominated by HARPER$++$ for the complete and ideal semantics.
Detailed results (ranking for each subtrack, with the cumulated runtime over the successfully solved instances) are given in Appendix~\ref{appendix:approximate}.

\begin{table}[htb]
\centering
\subfloat[Complete Semantics\label{tab:approximation-complete}]{
\begin{tabular}{ccc}
	\hline
	Rank & Solver & Score \\ \hline
	1 & HARPER++ & 747 \\
	2 & AFGCN & 668 \\
	\hline
\end{tabular}
}
\subfloat[Stable Semantics\label{tab:approximation-stable}]{
\begin{tabular}{ccc}
	\hline
	Rank & Solver & Score \\ \hline
	1 & AFGCN & 637 \\
	2 & HARPER++ & 457 \\
	\hline
\end{tabular}
}

\subfloat[Preferred Semantics\label{tab:approximation-preferred}]{
\begin{tabular}{ccc}
	\hline
	Rank & Solver & Score \\ \hline
	1 & AFGCN & 567 \\
	2 & HARPER++ & 438 \\
	\hline
\end{tabular}
}
\subfloat[Ideal Semantics\label{tab:approximation-ideal}]{
\begin{tabular}{cccc}
	\hline
	Rank & Solver & Score & Cumulated Runtime \\ \hline
	1 & HARPER++ & 108 & 9.848397 \\
	2 & AFGCN & 108 & 470.655630 \\
	\hline
\end{tabular}
}

\subfloat[Semi-Stable Semantics\label{tab:approximation-semi-stable}]{
\begin{tabular}{ccc}
	\hline
	Rank & Solver & Score \\ \hline
	1 & AFGCN & 522 \\
	2 & HARPER++ & 351 \\
	\hline
\end{tabular}
}
\subfloat[Stage Semantics\label{tab:approximation-stage}]{
\begin{tabular}{ccc}
	\hline
	Rank & Solver & Score \\ \hline
	1 & AFGCN & 392 \\
	2 & HARPER++ & 349 \\
	\hline
\end{tabular}
}
\caption{Rankings for the {\em Approximate track} \label{tab:ranking-approximate}}
\end{table}

\section{Conclusion}

This paper presents a short overview of the organization of ICCMA 2021 as well as the results of the competition. Detailed results and benchmark descriptions will be available soon at \url{http://argumentationcompetition.org/2021/index.html}.

Several ideas could be interesting for the organizers of the future editions of the competition. The first one consists in reviving the dynamic track that was introduced at ICCMA 2019. Unfortunately, the lack of participants for this track conducted us to remove it from ICCMA 2021, but it seems to us that real world application of argumentation requires techniques dedicated to dynamic scenarios, which makes it an important feature of next ICCMA editions. The same comment applies to structured argumentation, which was also removed from ICCMA 2021.
On the contrary, two interesting topics were introduced to ICCMA 2021 thanks to suggestions from the community. The first one is the track dedicated to approximate algorithms. For this first introduction of approximate algorithms at ICCMA, we chose to focus on the decision problems which are quite easy to evaluate. For the other reasoning tasks, it is not so easy to determine a relevant metric for evaluating the answer of an approximation algorithm. For instance, if the task is to determine a preferred extension of an AF, should we rank equally a solver that returns a (possibly non-maximal) admissible set and a solver that returns a conflicting set? This question, and similar ones, conducted us to exclude \CE-$\sigma$ and \SE-$\sigma$ for the approximate track. Metrics for evaluating approximate algorithms for these problems are thus an interesting question for the future of ICCMA.
The other question that arose during the organization of ICCMA 2021 is the issue of parallel computing. One solver ($\mu$-toksia) was submitted in two versions, one single-threaded and one multi-threaded. While we decided to run the multi-threaded versions without ranking it, a special track dedicated to parallel computing could be interesting in the future.

For concluding this short description of ICCMA 2021, we thank all the participants, as well as the ICCMA steering committee for providing some valuable feedback during the preliminary phases of the organization.
The competition was run on the CRIL computer cluster, that was funded by the French Ministry of Research and the {\em R\'egion Hauts de France} through CPER DATA.

\bibliographystyle{unsrt}  
\bibliography{references} 

\appendix

\section{Results of the Exact Track}\label{appendix:exact}
A multithreaded version of $\mu$-toksia has been submitted, and was run out of competition. Its results are presented here with those of single threaded solvers, although it was not eligible for an ICCMA award.

\subsection{Complete Semantics}
Table~\ref{tab:exact-complete} shows the results for the complete semantics.

\begin{table}[htb]
\centering
\subfloat[CE-CO\label{tab:exact-complete-CE}]{
\begin{tabular}{cccc}
	\hline
	Rank & Solver & Score & Runtime\\ \hline
1 & PYGLAF & 107 & 283014.802853 \\
2 & $\mu$-TOKSIA & 107 & 288298.145252 \\
3 & FUDGE & 107 & 288718.386884 \\
4 & ASPARTIX-V & 107 & 288844.420419 \\
5 & ConArg & 106 & 206539.159533 \\
n/a & $\mu$-TOKSIA-parallel & 105 & 287803.928120 \\
6 & A-Folio-DPDB & 80 & 224853.575249 \\
7 & MatrixX & 57 & 240888.182199 \\
	\hline
\end{tabular}
}
\subfloat[SE-CO\label{tab:exact-complete-SE}]{
\begin{tabular}{cccc}
	\hline
	Rank & Solver & Score & Runtime\\ \hline
1 & FUDGE & 587 & 931.517987 \\
2 & A-Folio-DPDB & 587 & 5706.033478 \\
3 & $\mu$-TOKSIA & 587 & 7540.231083 \\
n/a & $\mu$-TOKSIA-parallel & 587 & 12981.315567 \\
4 & MatrixX & 587 & 22954.152336 \\
5 & ASPARTIX-V & 587 & 30644.831330 \\
6 & PYGLAF & 586 & 56647.132293 \\
7 & ConArg & 107 & 199182.937802 \\
	\hline
\end{tabular}
}

\subfloat[DC-CO\label{tab:exact-complete-DC}]{
\begin{tabular}{cccc}
	\hline
	Rank & Solver & Score & Runtime\\ \hline
n/a & $\mu$-TOKSIA-parallel & 587 & 21217.410508 \\
1 & A-Folio-DPDB & 584 & 19690.986367 \\
2 & PYGLAF & 557 & 75189.484641 \\
3 & $\mu$-TOKSIA & 522 & 62550.500735 \\
4 &ASPARTIX-V & 506 & 107024.507674 \\
5 & FUDGE & 414 & 127037.443418 \\
6 & ConArg & 108 & 201311.092994 \\
7 & MatrixX & 58 & 240710.947344 \\
	\hline
\end{tabular}
}
\subfloat[DS-CO\label{tab:exact-complete-DS}]{
\begin{tabular}{cccc}
	\hline
	Rank & Solver & Score & Runtime\\ \hline
1 &FUDGE & 587 & 928.161398 \\
2 & A-Folio-DPDB & 587 & 6614.674043 \\
3 & $\mu$-TOKSIA & 587 & 8540.708610 \\
n/a & $\mu$-TOKSIA-parallel & 587 & 13997.941173 \\
4 & ASPARTIX-V & 587 & 30766.565703 \\
5 & PYGLAF & 585 & 56820.945178 \\
6 & ConArg & 107 & 199660.489916 \\
7 & MatrixX & 57 & 242088.917976 \\
	\hline
\end{tabular}
}

\subfloat[Overall Results for CO\label{tab:exact-complete-overall}]{
\begin{tabular}{cccc}
	\hline
	Rank & Solver & Score & Runtime\\ \hline
	n/a & $\mu$-TOKSIA-parallel & 1866 &  336000.595368\\
	1 & A-Folio DPDB &1838 & 256865.269137\\
	2 & PYGLAF &1835 & 471672.364965\\
	3 & $\mu$-toksia &1803 & 366929.58568\\
	4 & ASPARTIX-V21 & 1787 & 457280.325126\\
	5 & FUDGE & 1695 & 417615.509689\\
	6 & MatrixX & 759 & 746642.199855\\
	7 & ConArg & 428 & 806693.680245\\
	\hline
\end{tabular}
}
\caption{Result for the Exact Solvers on the Complete Semantics\label{tab:exact-complete}}
\end{table}

\subsection{Preferred Semantics}
Table~\ref{tab:exact-preferred} presents the results for the preferred semantics.

\begin{table}[htb]
\centering
\subfloat[CE-PR\label{tab:exact-preferred-CE}]{
\begin{tabular}{cccc}
	\hline
	Rank & Solver & Score & Runtime \\ \hline
1 & FUDGE & 107 & 284741.528303 \\
2 & $\mu$-TOKSIA & 107 & 285297.039682 \\
3 & ASPARTIX-V & 107 & 288914.836731 \\
4 & PYGLAF & 107 & 289391.458361 \\
5 & ConArg & 107 & 295524.859508 \\
n/a & $\mu$-TOKSIA-parallel & 105 & 287455.555046 \\
	\hline
\end{tabular}
}
\subfloat[SE-PR\label{tab:exact-preferred-SE}]{
\begin{tabular}{cccc}
	\hline
	Rank & Solver & Score & Runtime \\ \hline
1 & $\mu$-TOKSIA & 305 & 209598.856572 \\
2 & FUDGE & 298 & 206903.872679 \\
n/a & $\mu$-TOKSIA-parallel & 283 & 219758.945737 \\
3 & ASPARTIX-V & 266 & 231154.499826 \\
4 & PYGLAF & 210 & 255406.861456 \\
5 & ConArg & 107 & 291867.461181 \\
	\hline
\end{tabular}
}

\subfloat[DC-PR\label{tab:exact-preferred-DC}]{
\begin{tabular}{cccc}
	\hline
	Rank & Solver & Score & Runtime \\ \hline
n/a & $\mu$-TOKSIA-parallel & 587 & 21132.196079 \\
1 & PYGLAF & 557 & 75277.596631 \\
2 & $\mu$-TOKSIA & 523 & 62541.727318 \\
3 & ASPARTIX-V & 506 & 107016.876147 \\
4 & FUDGE & 414 & 127127.327070 \\
5 & ConArg & 108 & 199312.234514 \\
	\hline
\end{tabular}
}
\subfloat[DS-PR\label{tab:exact-preferred-DS}]{
\begin{tabular}{cccc}
	\hline
	Rank & Solver & Score & Runtime \\ \hline
1 & PYGLAF & 425 & 143255.632741 \\
2 & FUDGE & 371 & 159974.358059 \\
3 & $\mu$-TOKSIA & 275 & 223662.448184 \\
n/a & $\mu$-TOKSIA-parallel & 220 & 242186.574490 \\
4 & ASPARTIX-V & 173 & 265994.171780 \\
5 & ConArg & 107 & 292285.817332 \\
	\hline
\end{tabular}
}

\subfloat[Overall Results for PR\label{tab:exact-preferred-overall}]{
\begin{tabular}{cccc}
	\hline
	Rank & Solver & Score & Runtime \\ \hline
	1 & PYGLAF& 1299 & 763331.549189\\
	2 & $\mu$-toksia & 1210 & 781100.071756\\
	n/a & $\mu$-TOKSIA-parallel & 1195 & 770533.271352\\
	3 & FUDGE & 1190 & 778747.086111\\
	4 & ASPARTIX-V21 & 1052 & 893080.384484\\
	5 & ConArg & 429 & 1078990.372535\\ 
	\hline
\end{tabular}
}
\caption{Result for the Exact Solvers on the Preferred Semantics\label{tab:exact-preferred}}
\end{table}

\subsection{Semi-Stable Semantics}
Table~\ref{tab:exact-semi-stable} presents results for the semi-stable semantics.

\begin{table}[htb]
\centering
\subfloat[CE-SST\label{tab:exact-semi-stable-CE}]{
\begin{tabular}{cccc}
	\hline
	Rank & Solver & Score & Runtime \\ \hline
1 & ConArg & 107 & 201218.617540 \\
2 & ASPARTIX-V & 107 & 288935.416848 \\
3 & $\mu$-TOKSIA & 107 & 289033.884355 \\
4 & PYGLAF & 106 & 228421.036070 \\
n/a & $\mu$-TOKSIA-parallel & 105 & 291565.135040 \\
	\hline
\end{tabular}
}
\subfloat[SE-SST\label{tab:exact-semi-stable-SE}]{
\begin{tabular}{cccc}
	\hline
	Rank & Solver & Score & Runtime \\ \hline
1 & PYGLAF & 442 & 149977.311142 \\
2 & $\mu$-TOKSIA & 285 & 222654.386562 \\
n/a & $\mu$-TOKSIA-parallel & 271 & 236882.931798 \\
3 & ASPARTIX-V & 215 & 245358.997303 \\
4 & ConArg & 107 & 198299.550176 \\
	\hline
\end{tabular}
}

\subfloat[DC-SST\label{tab:exact-semi-stable-DC}]{
\begin{tabular}{cccc}
	\hline
	Rank & Solver & Score & Runtime \\ \hline
1 & PYGLAF & 485 & 114847.031276 \\
2 & $\mu$-TOKSIA & 481 & 83988.743044 \\
n/a & $\mu$-TOKSIA-parallel & 476 & 90649.801238 \\
3 & ASPARTIX-V & 210 & 250516.541457 \\
4 & ConArg & 107 & 200332.750174 \\
	\hline
\end{tabular}
}
\subfloat[DS-SST\label{tab:exact-semi-stable-DS}]{
\begin{tabular}{cccc}
	\hline
	Rank & Solver & Score &  Runtime\\ \hline
1 & PYGLAF & 482 & 115633.357949 \\
2 & $\mu$-TOKSIA & 230 & 242028.695258 \\
3 & ASPARTIX-V & 212 & 246004.720807 \\
n/a & $\mu$-TOKSIA-parallel & 156 & 272773.151090 \\
4 & ConArg & 107 & 199271.215080 \\
	\hline
\end{tabular}
}

\subfloat[Overall Results for SST\label{tab:exact-semi-stable-overall}]{
\begin{tabular}{cccc}
	\hline
	Rank & Solver & Score &  Runtime\\ \hline
	1 & PYGLAF & 1515 & 608878.736437\\
	n/a & $\mu$-TOKSIA-parallel & 1008 &  891871.019166\\
	2 & $\mu$-toksia & 1103 & 837705.709219\\
	3 & ASPARTIX-V21 &  744 & 1030815.676415\\
	4 & ConArg & 428 & 799122.13297\\
	\hline
\end{tabular}
}
\caption{Results for the Exact Solvers on the Semi-Stable Semantics \label{tab:exact-semi-stable}}
\end{table}

\subsection{Results for the Stable Semantics}
Table~\ref{tab:exact-stable} presents results for the stable semantics.

\begin{table}[htb]
\centering
\subfloat[CE-ST\label{tab:exact-stable-CE}]{
\begin{tabular}{cccc}
	\hline
	Rank & Solver & Score & Runtime \\ \hline
1 & PYGLAF & 176 & 262076.166538 \\
2 & FUDGE & 171 & 261123.826755 \\
3 & $\mu$-TOKSIA & 164 & 263995.703902 \\
3 & ASPARTIX-V & 163 & 266764.981322 \\
n/a & $\mu$-TOKSIA-parallel & 159 & 272437.936664 \\
4 & A-Folio-DPDB & 125 & 247167.250461 \\
5 & ConArg & 107 & 143824.774571 \\
6 & MatrixX & 62 & 248857.004617 \\
	\hline
\end{tabular}
}
\subfloat[SE-ST\label{tab:exact-stable-SE}]{
\begin{tabular}{cccc}
	\hline
	Rank & Solver & Score & Runtime\\ \hline
1 & A-Folio-DPDB & 577 & 57664.945319 \\
2 & PYGLAF & 508 & 116669.565521 \\
3 & FUDGE & 457 & 135396.148397 \\
4 & ASPARTIX-V & 399 & 172533.607683 \\
5 & $\mu$-TOKSIA & 387 & 167510.314348 \\
n/a & $\mu$-TOKSIA-parallel & 371 & 190641.447195 \\
6 & ConArg & 107 & 141595.187820 \\
7 & MatrixX & 71 & 243684.820216 \\
	\hline
\end{tabular}
}

\subfloat[DC-ST\label{tab:exact-stable-DC}]{
\begin{tabular}{cccc}
	\hline
	Rank & Solver & Score & Runtime \\ \hline
1 & A-Folio-DPDB & 585 & 35566.107761 \\
2 & PYGLAF & 548 & 89570.851193 \\
3 & FUDGE & 505 & 92693.283381 \\
4 & $\mu$-TOKSIA & 504 & 94875.690027 \\
5 & ASPARTIX-V & 472 & 118986.052239 \\
n/a & $\mu$-TOKSIA-parallel & 463 & 126633.414211 \\
6 & ConArg & 108 & 142981.220887 \\
7 & MatrixX & 64 & 247636.605310 \\
	\hline
\end{tabular}
}
\subfloat[DS-ST\label{tab:exact-stable-DS}]{
\begin{tabular}{cccc}
	\hline
	Rank & Solver & Score & \\ \hline
1 & A-Folio-DPDB & 575 & 58912.571800 \\
2 & PYGLAF & 511 & 115965.465025 \\
3 & FUDGE & 452 & 139597.720683 \\
4 & ASPARTIX-V & 395 & 172828.848761 \\
5 & $\mu$-TOKSIA & 386 & 166901.037702 \\
n/a & $\mu$-TOKSIA-parallel & 373 & 189986.356391 \\
6 & ConArg & 107 & 141538.881464 \\
7 & MatrixX & 62 & 247830.428679 \\
	\hline
\end{tabular}
}

\subfloat[Overall Results for ST\label{tab:exact-stable-overall}]{
\begin{tabular}{cccc}
	\hline
	Rank & Solver & Score & Runtime \\ \hline
	1 & A-Folio-DPDB & 1862 & 399310.875341 \\
	2 & PYGLAF & 1743 & 584282.048277\\
	3 & FUDGE & 1585 & 628810.979216\\
	4 & $\mu$-toksia & 1441 & 693282.745979 \\
	5 & ASPARTIX-V21 & 1429 & 713113.490005\\
	n/a & $\mu$-TOKSIA-parallel & 1366 & 779699.154461\\
	6 & ConArg & 429 & 569940.064742\\
	7 & MatrixX & 259 & 988008.858822 \\
	\hline
\end{tabular}
}
\caption{Results for the Exact Solvers on the Semi-Stable Semantics \label{tab:exact-stable}}
\end{table}

\subsection{Stage Semantics}
Table~\ref{tab:exact-stage} shows the results for the stage semantics. PYGLAF was removed from the ranking because of wrong results on CE-STG.

\begin{table}[htb]
\centering
\subfloat[CE-STG\label{tab:exact-stage-CE}]{
\begin{tabular}{cccc}
	\hline
	Rank & Solver & Score & Runtime \\ \hline
1 & $\mu$-TOKSIA & 107 & 287359.108885 \\
2 & ASPARTIX-V & 107 & 288840.295997 \\
3 & ConArg & 105 & 81427.551066 \\
n/a & $\mu$-TOKSIA-parallel & 105 & 290706.406160 \\
	\hline
\end{tabular}
}
\subfloat[SE-STG\label{tab:exact-stage-SE}]{
\begin{tabular}{cccc}
	\hline
	Rank & Solver & Score & Runtime \\ \hline
1 & PYGLAF & 504 & 112541.710700 \\
2 & ASPARTIX-V & 271 & 225125.628443 \\
3 & $\mu$-TOKSIA & 236 & 236304.243330 \\
n/a & $\mu$-TOKSIA-parallel & 180 & 262418.192826 \\
4 & ConArg & 107 & 79497.777484 \\
	\hline
\end{tabular}
}

\subfloat[DC-STG\label{tab:exact-stage-DC}]{
\begin{tabular}{cccc}
	\hline
	Rank & Solver & Score & Runtime \\ \hline
1 & ASPARTIX-V & 245 & 235698.489668 \\
2 & $\mu$-TOKSIA & 219 & 242756.715179 \\
n/a & $\mu$-TOKSIA-parallel & 166 & 267034.642346 \\
3 & PYGLAF & 149 & 276541.935306 \\
4 & ConArg & 106 & 81097.743104 \\
	\hline
\end{tabular}
}
\subfloat[DS-STG\label{tab:exact-stage-DS}]{
\begin{tabular}{cccc}
	\hline
	Rank & Solver & Score & Runtime \\ \hline
1 & ASPARTIX-V & 256 & 230874.248362 \\
2 & $\mu$-TOKSIA & 226 & 240280.159325 \\
n/a & $\mu$-TOKSIA-parallel & 176 & 265249.960730 \\
3 & PYGLAF & 164 & 270734.066210 \\
4 & ConArg & 107 & 79347.232359 \\
	\hline
\end{tabular}
}

\subfloat[Overall Results for STG\label{tab:exact-stage-overall}]{
\begin{tabular}{cccc}
	\hline
	Rank & Solver & Score & Runtime \\ \hline
	1 & ASPARTIX-V21 & 879 & 980538.66247\\
	2 & $\mu$-toksia & 788 & 1006700.226719\\
	n/a & $\mu$-TOKSIA-parallel & 627 &  1085409.202062\\
	3 & ConArg & 425 & 321370.304013\\
	\hline
\end{tabular}
}
\caption{Results for the Exact Solvers on the Stage Semantics\label{tab:exact-stage}}
\end{table}

\subsection{Ideal Semantics}
Table~\ref{tab:exact-ideal} shows the results for the ideal semantics.

\begin{table}[htb]
\centering
\subfloat[SE-ID\label{tab:exact-ideal-SE}]{
\begin{tabular}{cccc}
	\hline
	Rank & Solver & Score & Runtime \\ \hline
1 & FUDGE & 234 & 242189.672977 \\
n/a & $\mu$-TOKSIA-parallel & 149 & 279337.091744 \\
2 & PYGLAF & 118 & 291221.721682 \\
3 & $\mu$-TOKSIA & 108 & 287179.393261 \\
4 & ConArg & 107 & 300671.098342 \\
5 & ASPARTIX-V & 104 & 279290.787518 \\
	\hline
\end{tabular}
}
\subfloat[DS-ID\label{tab:exact-ideal-DS}]{
\begin{tabular}{cccc}
	\hline
	Rank & Solver & Score & Runtime \\ \hline
1 & FUDGE & 258 & 222757.487117 \\
2 & ASPARTIX-V & 202 & 263712.239254 \\
n/a & $\mu$-TOKSIA-parallel & 151 & 278761.114962 \\
3 & PYGLAF & 120 & 290737.838592 \\
4 & $\mu$-TOKSIA & 108 & 287582.446341 \\
5 & ConArg & 107 & 300511.844717 \\
	\hline
\end{tabular}
}

\subfloat[Overall Results for ID\label{tab:exact-ideal-overall}]{
\begin{tabular}{cccc}
	\hline
	Rank & Solver & Score & Runtime \\ \hline
	1 & FUDGE & 492 & 464947.160094\\
	2 & ASPARTIX-V21 & 306 & 543003.026772 \\
	n/a & $\mu$-TOKSIA-parallel & 300 &  558098.206706\\
	3 &  PYGLAF & 238 & 581959.560274\\
	4 & $\mu$-toksia & 216&  574761.838951\\
	5 & ConArg & 214 & 601182.943059\\
	\hline
\end{tabular}
}
\caption{Results for the Exact Solvers on the Ideal Semantics\label{tab:exact-ideal}}
\end{table}

\section{Results of the Approximate Track}\label{appendix:approximate}
\subsection{Complete Semantics}
Table~\ref{tab:approximation-complete} shows the results for the complete semantics.

\begin{table}[htb]
\centering
\subfloat[DC-CO\label{tab:approximation-complete-DC}]{
\begin{tabular}{cccc}
	\hline
	Rank & Solver & Score & Runtime \\ \hline
1 & AFGCN & 291 & 15273.326580 \\
2 & HARPER++ & 160 & 677.187391 \\
	\hline
\end{tabular}
}
\subfloat[DS-CO\label{tab:approximation-complete-DS}]{
\begin{tabular}{cccc}
	\hline
	Rank & Solver & Score & Runtime \\ \hline
1 & HARPER++ & 587 & 958.272555 \\
2 & AFGCN & 377 & 18882.239960 \\
	\hline
\end{tabular}
}

\subfloat[Overall Results for CO\label{tab:approximation-complete-overall}]{
\begin{tabular}{cccc}
	\hline
	Rank & Solver & Score & Runtime \\ \hline
	1 & HARPER++ & 747 & 1635.459946\\
	2 & AFGCN & 668 & 34155.56654\\
	\hline
\end{tabular}
}
\caption{Results for the Approximate Solvers on the Complete Semantics \label{tab:approximation-complete}}
\end{table}

\subsection{Preferred Semantics}
Table~\ref{tab:approximation-preferred} shows the results for the preferred semantics.

\begin{table}[htb]
\centering
\subfloat[DC-PR\label{tab:approximation-preferred-DC}]{
\begin{tabular}{cccc}
	\hline
	Rank & Solver & Score & Runtime \\ \hline
1 & AFGCN & 288 & 15253.015020 \\
2 & HARPER++ & 159 & 674.846465 \\
	\hline
\end{tabular}
}
\subfloat[DS-PR\label{tab:approximation-preferred-DS}]{
\begin{tabular}{cccc}
	\hline
	Rank & Solver & Score & Runtime \\ \hline
1 & HARPER++ & 279 & 68.114114 \\
2 & AFGCN & 279 & 3138.994710 \\
	\hline
\end{tabular}
}

\subfloat[Overall Results for PR\label{tab:approximation-preferred-overall}]{
\begin{tabular}{cccc}
	\hline
	Rank & Solver & Score & Runtime \\ \hline
	1 & AFGCN & 567 & 18392.00973\\
	2 & HARPER++ & 438 & 742.960579\\
	\hline
\end{tabular}
}
\caption{Results for the Approximate Solvers on the Preferred Semantics \label{tab:approximation-preferred}}
\end{table}

\subsection{Semi-Stable Semantics}
Table~\ref{tab:approximation-semi-stable} shows the results for the semi-stable semantics.

\begin{table}[htb]
\centering
\subfloat[DC-SST\label{tab:approximation-semi-stable-DC}]{
\begin{tabular}{cccc}
	\hline
	Rank & Solver & Score & Runtime \\ \hline
1 & AFGCN & 290 & 13470.044680 \\
2 & HARPER++ & 119 & 621.545732 \\
	\hline
\end{tabular}
}
\subfloat[DS-SST\label{tab:approximation-semi-stable-DS}]{
\begin{tabular}{cccc}
	\hline
	Rank & Solver & Score & Runtime \\ \hline
1 & HARPER++ & 232 & 50.005328 \\
2 & AFGCN & 232 & 2292.188910 \\
	\hline
\end{tabular}
}

\subfloat[Overall Results for SST\label{tab:approximation-semi-stable-overall}]{
\begin{tabular}{cccc}
	\hline
	Rank & Solver & Score & Runtime \\ \hline
	1 & AFGCN & 522 & 15762.23359\\
	2 & HARPER++ & 351 & 671.55106\\
	\hline
\end{tabular}
}
\caption{Results for the Approximate Solvers on the Semi-Stable Semantics \label{tab:approximation-semi-stable}}
\end{table}

\subsection{Stable Semantics}
Table~\ref{tab:approximation-stable} shows the results for the stable semantics.

\begin{table}[htb]
\centering
\subfloat[DC-ST\label{tab:approximation-stable-DC}]{
\begin{tabular}{cccc}
	\hline
	Rank & Solver & Score & Runtime\\ \hline
1 & AFGCN & 300 & 14610.356710 \\
2 & HARPER++ & 142 & 642.296641 \\
	\hline
\end{tabular}
}
\subfloat[DS-ST\label{tab:approximation-stable-DS}]{
\begin{tabular}{cccc}
	\hline
	Rank & Solver & Score & Runtime\\ \hline
1 & HARPER++ & 337 & 289.741214 \\
2 & AFGCN & 315 & 9290.415250 \\
	\hline
\end{tabular}
}

\subfloat[Overall Results for ST\label{tab:approximation-stable-overall}]{
\begin{tabular}{cccc}
	\hline
	Rank & Solver & Score & Runtime\\ \hline
	1 & AFGCN & 637 & 23900.77196\\
	2 & HARPER++ & 457 & 932.037855\\
	\hline
\end{tabular}
}
\caption{Results for the Approximate Solvers on the Stable Semantics \label{tab:approximation-stable}}
\end{table}

\subsection{Stage Semantics}
Table~\ref{tab:approximation-stage} shows the results for the stage semantics.

\begin{table}[htb]
\centering
\subfloat[DC-STG\label{tab:approximation-stage-DC}]{
\begin{tabular}{cccc}
	\hline
	Rank & Solver & Score & Runtime \\ \hline
1 & AFGCN & 164 & 2228.788830 \\
2 & HARPER++ & 121 & 48.270998 \\
	\hline
\end{tabular}
}
\subfloat[DS-STG\label{tab:approximation-stage-DS}]{
\begin{tabular}{cccc}
	\hline
	Rank & Solver & Score & Runtime \\ \hline
1 & HARPER++ & 228 & 49.946634 \\
2 & AFGCN & 228 & 2301.554530 \\
	\hline
\end{tabular}
}

\subfloat[Overall Results for STG\label{tab:approximation-stage-overall}]{
\begin{tabular}{cccc}
	\hline
	Rank & Solver & Score & Runtime \\ \hline
	1 & AFGCN & 392 & 4530.34336\\
	2 & HARPER++ & 349 & 98.217632\\
	\hline
\end{tabular}
}
\caption{Results for the Approximate Solvers on the Stage Semantics \label{tab:approximation-stage}}
\end{table}

\subsection{Ideal Semantics}
Table~\ref{tab:approximation-ideal} shows the results for the ideal semantics.

\begin{table}[htb]
\centering
\begin{tabular}{cccc}
	\hline
	Rank & Solver & Score & Cumulated Runtime \\ \hline
	1 & HARPER++ & 108 & 9.848397 \\
	2 & AFGCN & 108 & 470.655630 \\
	\hline
\end{tabular}
\caption{Results for the Approximate Solvers on the Ideal Semantics (DS-ID) \label{tab:approximation-ideal}}
\end{table}

%
\end{document}